%% file: main.tex
\documentclass[10pt, twocolumn, letterpaper]{article}

\usepackage{cvpr}
\usepackage{times}
\usepackage{epsfig}
\usepackage{graphicx}
\usepackage{amsmath}
\usepackage{amssymb}
\usepackage{multirow}
\usepackage[numbers,sort&compress]{natbib}
\usepackage[table,xcdraw]{xcolor}
\usepackage[ruled]{algorithm2e}
\usepackage{dsfont}
\usepackage[caption=false]{subfig}
\usepackage[flushleft]{threeparttable}
\usepackage{nopageno}
\usepackage{url}
\usepackage{cleveref}

\DeclareMathOperator*{\argmin}{arg\,min} 

 \cvprfinalcopy 

\def\cvprPaperID{2417} 

\title
{
    Designing Energy-Efficient Convolutional Neural Networks \\
    using Energy-Aware Pruning
}

\author
{
    Tien-Ju Yang, Yu-Hsin Chen, Vivienne Sze\\
    Massachusetts Institute of Technology\\
    {\tt\small \{tjy, yhchen, sze\}@mit.edu}
}

\begin{document}
\maketitle


\input{0_abstract}
\vspace{-5pt}


\input{1_introduction}

\input{2_energy_evaluation}

\input{3_related_work}

\input{4_energy_aware_pruning}

\input{5_experiment_results}

\input{6_conclusion}


\clearpage
{\small
\bibliographystyle{ieeetr}
\bibliography{__references}
}


\end{document}

%% file: 0_abstract.tex
\begin{abstract}
\vspace{-5pt}
Deep convolutional neural networks (CNNs) are indispensable to state-of-the-art computer vision algorithms. However, they are still rarely deployed on battery-powered mobile devices, such as smartphones and wearable gadgets, where vision algorithms can enable many revolutionary real-world applications. The key limiting factor is the high energy consumption of CNN processing due to its high computational complexity. While there are many previous efforts that try to reduce the CNN model size or the amount of computation, we find that they do not necessarily result in lower energy consumption. Therefore, these targets do not serve as a good metric for energy cost estimation.

To close the gap between CNN design and energy consumption optimization, we propose an energy-aware pruning algorithm for CNNs that directly uses the energy consumption of a CNN to guide the pruning process. The energy estimation methodology uses parameters extrapolated from actual hardware measurements. The proposed layer-by-layer pruning algorithm also prunes more aggressively than previously proposed pruning methods by minimizing the error in the output feature maps instead of the filter weights. For each layer, the weights are first pruned and then locally fine-tuned with a closed-form least-square solution to quickly restore the accuracy. After all layers are pruned, the entire network is globally fine-tuned using back-propagation. With the proposed pruning method, the energy consumption of AlexNet and GoogLeNet is reduced by 3.7$\times$ and 1.6$\times$, respectively, with less than 1\% top-5 accuracy loss. We also show that reducing the number of target classes in AlexNet greatly decreases the number of weights, but has a limited impact on energy consumption.

\end{abstract}

%% file: 1_introduction.tex
\vspace{-10pt}
\section{Introduction}
\label{sec:introduction}

In recent years, deep convolutional neural networks (CNNs) have become the state-of-the-art solution for many computer vision applications and are ripe for real-world deployment~\cite{nature2015-lecun-deep_learning}. However, CNN processing incurs high energy consumption due to its high computational complexity~\cite{nvidia2015_deep_learning_performance}. As a result, battery-powered devices still cannot afford to run state-of-the-art CNNs due to their limited energy budget. For example, smartphones nowadays cannot even run object classification with AlexNet~\cite{nips2012-krizhevsky-alexnet} in real-time for more than an hour. Hence, energy consumption has become the primary issue of bridging CNNs into practical computer vision applications.

In addition to accuracy, the design of modern CNNs is starting to incorporate new metrics to make it more favorable in real-world environments. For example, the trend is to simultaneously reduce the overall CNN model size and/or simplify the computation while going deeper. This is achieved either by pruning the weights of existing CNNs, \ie, making the filters sparse by setting some of the weights to zero~\cite{nips1990-lecun-opt_brain_damage, nips1993-hassibi-opt_brain_surgeon, book-hertz-theory-neural-computation, nips2015-han-learn_conn, iclr2016-han-deep_comp, arxiv2016-jin-skinny_net, nips2016-guo-dynamic_surgery, tnn1993-reed-pruning_survey, arxiv2016-hu-remove_zero_freq, bmvc2015-srinivas-data_free_prune, iclr2016-mariet-diversity-networks}, or by designing new CNNs with (1) highly bitwidth-reduced weights and operations (\eg, XNOR-Net and BWN~\cite{eccv2016-rastegari-xnor_net}) or (2) compact layers with fewer weights (\eg, Network-in-Network~\cite{iclr2013-lin-nin}, GoogLeNet~\cite{cvpr2015-szegedy-googlenet}, SqueezeNet~\cite{arxiv2016-iandola-squeezenet}, and ResNet~\cite{cvpr2016-he-resnet}).

However, neither the number of weights nor the number of operations in a CNN directly reflect its actual energy consumption. A CNN with a smaller model size or fewer operations can still have higher overall energy consumption. This is because the sources of energy consumption in a CNN consist of not only computation but also memory accesses. In fact, fetching data from the DRAM for an operation consumes \emph{orders of magnitude higher energy} than the computation itself~\cite{isscc2014-horowitz-energy_problem}, and the energy consumption of a CNN is dominated by memory accesses for both filter weights and \emph{feature maps}. The total number of memory accesses is a function of the CNN shape configuration~\cite{isca2016-chen-eyeriss_dataflow} (\ie, filter size, feature map resolution, number of channels, and number of filters); different shape configurations can lead to different amounts of memory accesses, and thus energy consumption, even under the same number of weights or operations. Therefore, there is still no evidence showing that the aforementioned approaches can directly optimize the energy consumption of a CNN. In addition, there is currently no way for researchers to estimate the energy consumption of a CNN at design time.

The key to closing the gap between CNN design and energy efficiency optimization is to directly use energy, instead of the number of weights or operations, as a metric to guide the design. In order to obtain realistic estimate of energy consumption at design time of the CNN, we use the framework proposed in~\cite{isca2016-chen-eyeriss_dataflow} that models the two sources of energy consumption in a CNN (computation and memory accesses), and use energy numbers extrapolated from actual hardware measurements~\cite{isscc2016-chen-eyeriss_chip}. We then extend it to further model the impact of data sparsity and bitwidth reduction. The setup targets battery-powered platforms, such as smartphones and wearable devices, where hardware resources (\ie, computation and memory) are limited and energy efficiency is of utmost importance.

We further propose a new CNN pruning algorithm with the goal to minimize overall energy consumption with marginal accuracy degradation. Unlike the previous pruning methods, it directly minimizes the changes to the output feature maps as opposed to the changes to the filters and achieves a higher compression ratio (\ie, the number of removed weights divided by the number of total weights). With the ability to directly estimate the energy consumption of a CNN, the proposed pruning method identifies the parts of a CNN where pruning can maximally reduce the energy cost, and prunes the weights more aggressively than previously proposed methods to maximize the energy reduction.


In summary, the key contributions of this work include:
\begin{list}{\labelitemi}{\leftmargin=1em}
  \setlength{\topmargin}{0pt}
  \setlength{\itemsep}{0em}
  \setlength{\parskip}{0pt}
  \setlength{\parsep}{0pt}
  
\item \textbf{Energy Estimation Methodology}: Since the number of weights or operations does not necessarily serve as a good metric to guide the CNN design toward higher energy efficiency, we directly use the energy consumption of a CNN to guide its design. This methodology is based on the framework proposed in~\cite{isca2016-chen-eyeriss_dataflow} for realistic battery-powered systems, \eg, smartphones, wearable devices, etc. We then further extend it to model the impact of data sparsity and bitwidth reduction. The corresponding energy estimation tool is available at~\cite{website2017-yang-energy-estimation}.

\item \textbf{Energy-Aware Pruning}: We propose a new layer-by-layer pruning method that can aggressively reduce the number of non-zero weights by minimizing changes in feature maps as opposed to changes in filters. To maximize the energy reduction, the algorithm starts pruning the layers that consume the most energy instead of with the largest number of weights, since pruning becomes more difficult as more layers are pruned. Each layer is first pruned and the preserved weights are locally fine-tuned with a closed-form least-square solution to quickly restore the accuracy and increase the compression ratio. After all the layers are pruned, the entire network is further globally fine-tuned by back-propagation. As a result, for AlexNet, we can reduce energy consumption by 3.7$\times$ after pruning, which is 1.7$\times$ lower than pruning with the popular network pruning method proposed in~\cite{iclr2016-han-deep_comp}. Even for a compact CNN, such as GoogLeNet, the proposed pruning method can still reduce energy consumption by 1.6$\times$. The pruned models will be released at~\cite{website2017-yang-energy-estimation}. As many embedded applications only require a limited set of classes, we also show the impact of pruning AlexNet for a reduced number of target classes.


\item \textbf{Energy Consumption Analysis of CNNs}: We evaluate the energy versus accuracy trade-off of widely-used or pruned CNN models. Our key insights are that (1) maximally reducing weights or the number of MACs in a CNN does not necessarily result in optimized energy consumption, and \emph{feature maps need to be factored in}, (2) convolutional (CONV) layers, instead of fully-connected (FC) layers, dominate the overall energy consumption in a CNN, (3) deeper CNNs with fewer weights, \eg, GoogLeNet and SqueezeNet, do not necessarily consume less energy than shallower CNNs with more weights, \eg, AlexNet, and (4) sparsifying the filters can provide equal or more energy reduction than reducing the bitwidth (even to binary) of weights.

\end{list}

%% file: 2_energy_evaluation.tex
\section{Energy Estimation Methodology}
\label{sec:energy_estimation}

\subsection{Background and Motivation}
\label{subsec:energy_estimation_background}

\begin{figure*}[t]
    \begin{center}
        \includegraphics[width=0.85\linewidth]{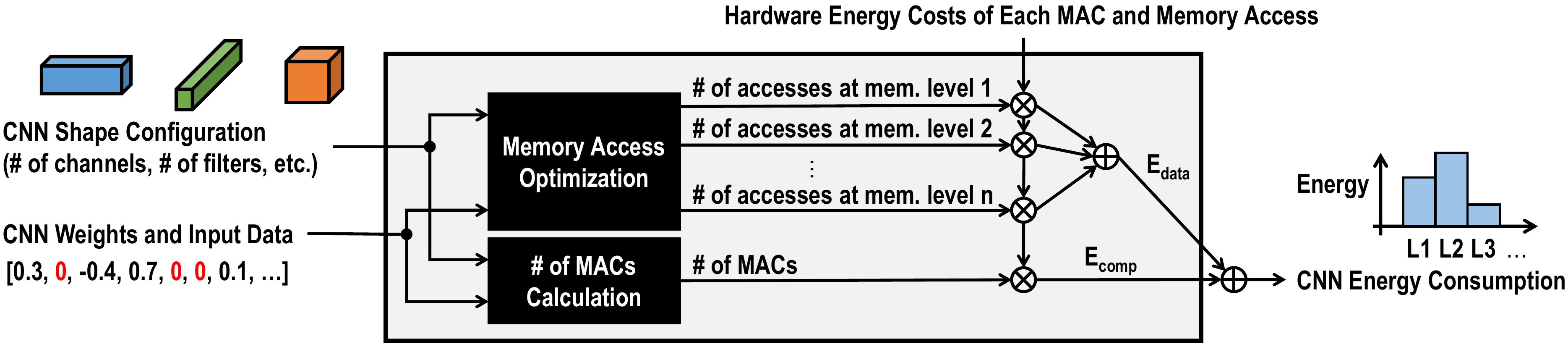}
        \caption{    The energy estimation methodology is based on the framework proposed in~\cite{isca2016-chen-eyeriss_dataflow}, which optimizes the memory accesses at each level of the memory hierarchy to achieve the lowest energy consumption. We then further account for the impact of data sparsity and bitwidth reduction, and use energy numbers extrapolated from actual hardware measurements of~\cite{isscc2016-chen-eyeriss_chip} to calculate the energy for both computation and data movement.
                }
                \vspace{-20pt}
        \label{fig:energy_estimation_methodology}
    \end{center}
\end{figure*}

Multiply-and-accumulate (MAC) operations in CONV and FC layers account for over 99\% of total operations in state-of-the-art CNNs~\cite{nips2012-krizhevsky-alexnet, iclr2015-simonyan-vgg, cvpr2015-szegedy-googlenet, cvpr2016-he-resnet}, and therefore dominate both processing runtime and energy consumption. The energy consumption of MACs comes from computation and memory accesses for the required data, including both weights and feature maps. While the amount of computation increases linearly with the number of MACs, the amount of required data does not necessarily scale accordingly due to data reuse, \ie, the same data value is used for multiple MACs. This implies that some data have a higher impact on energy than others, since they are accessed more often. In other words, removing the data that are reused more has the potential to yield higher energy reduction.

Data reuse in a CNN arises in many ways, and is determined by the shape configurations of different layers. In CONV layers, due to its weight sharing property, each weight and input activation are reused many times according to the resolution of output feature maps and the size of filters, respectively. In both CONV and FC layers, each input activation is also reused across all filters for different output channels within the same layer. When input batching is applied, each weight is further reused across all input feature maps in both types of layers. Overall, CONV layers usually present much more data reuse than FC layers. Therefore, as a general rule of thumb, each weight and activation in CONV layers have a higher impact on energy than in FC layers.

While data reuse serves as a good metric for comparing relative energy impact of data, it does not directly translate to the actual energy consumption. This is because modern hardware processors implement multiple levels of memory hierarchy, \eg, DRAM and multi-level buffers, to amortize the energy cost of memory accesses. The goal is to access data more from the less energy-consuming memory levels, which usually have less storage capacity, and thus minimize data accesses to the more energy-consuming memory levels. Therefore, the total energy cost to access a single piece of data with many reuses can vary a lot depending on how the accesses spread across different memory levels, and minimizing overall energy consumption using the memory hierarchy is the key to energy-efficient processing of CNNs.

\subsection{Methodology}

With the idea of exploiting data reuse in a multi-level memory hierarchy, Chen et al.~\cite{isca2016-chen-eyeriss_dataflow} have presented a framework that can estimate the energy consumption of a CNN for inference. As shown in Fig~\ref{fig:energy_estimation_methodology}, for each CNN layer, the framework calculates the energy consumption by dividing it into two parts: computation energy consumption, $E_{comp}$, and data movement energy consumption, $E_{data}$. $E_{comp}$ is calculated by counting the number of MACs in the layer and weighing it with the energy consumed by running each MAC operation in the computation core. $E_{data}$ is calculated by counting the number of memory accesses at each level of the memory hierarchy in the hardware and weighing it with the energy consumed by each access of that memory level. To obtain the number of memory accesses,~\cite{isca2016-chen-eyeriss_dataflow} proposes an optimization procedure to search for the optimal number of accesses for all data types (feature maps and weights) at all levels of memory hierarchy that results in the lowest energy consumption. For energy numbers of each MAC operation and memory access, we use numbers extrapolated from actual hardware measurements of the platform targeting battery-powered devices~\cite{isscc2016-chen-eyeriss_chip}.

Based on the aforementioned framework, we have created a methodology that further accounts for the impact of data sparsity and bitwidth reduction on energy consumption. For example, we assume that the computation of a MAC and its associated memory accesses can be skipped completely when either of its input activation or weight is zero. Lossless data compression is also applied on the sparse data to save the cost of both on-chip and off-chip data movement. The impact of bitwidth is quantified by scaling the energy cost of different hardware components accordingly. For instance, the energy consumption of a multiplier scales with the bitwidth quadratically, while that of a memory access only scales its energy linearly.

\subsection{Potential Impact}

With this methodology, we can quantify the difference in energy costs between various popular CNN models and methods, such as increasing data sparsity or aggressive bitwidth reduction (discussed in Sec.~\ref{sec:experiment_results}). More importantly, it provides a gateway for researchers to assess the energy consumption of CNNs at design time, which can be used as a feedback that leads to CNN designs with significantly reduced energy consumption. In Sec.~\ref{sec:energy_aware_pruning}, we will describe an energy-aware pruning method that uses the proposed energy estimation method for deciding the layer pruning priority.


%% file: 3_related_work.tex
\section{CNN Pruning: Related Work}
\label{sec:related_work}

\textbf{Weight pruning}. There is a large body of work that aims to reduce the CNN model size by pruning weights while maintaining accuracy. LeCun et al.~\cite{nips1990-lecun-opt_brain_damage} and Hassibi et al.~\cite{nips1993-hassibi-opt_brain_surgeon} remove the weights based on the sensitivity of the final objective function to that weight (\ie, remove the weights with the least sensitivity first). However, the complexity of computing the sensitivity is too high for large networks, so the magnitude-based pruning methods~\cite{book-hertz-theory-neural-computation} use the magnitude of a weight to approximate its sensitivity; specifically, the small-magnitude weights are removed first. Han et al.~\cite{nips2015-han-learn_conn, iclr2016-han-deep_comp} applied this idea to recent networks and achieved large model size reduction. They iteratively prune and globally fine-tune the network, and the pruned weights will always be zero after being pruned. Jin et al.~\cite{arxiv2016-jin-skinny_net} and Guo et al.~\cite{nips2016-guo-dynamic_surgery} extend the magnitude-based methods to allow the restoration of the pruned weights in the previous iterations, with tightly coupled pruning and global fine-tuning stages, for greater model compression. However, all the above methods evaluate whether to prune each weight independently and do not account for correlation between weights~\cite{tnn1993-reed-pruning_survey}.  When the compression ratio is large, the aggregate impact of many weights can have a large impact on the output; thus, failing to consider the combined influence of the weights on the output limits the achievable compression ratio.

\textbf{Filter pruning}. Rather than investigating the removal of each individual weight (fine-grained pruning), there is also work that investigates removing entire filters (coarse-grained pruning). Hu et al.~\cite{arxiv2016-hu-remove_zero_freq} proposed removing  filters that frequently generate zero outputs after the ReLU layer in the validation set. Srinivas et al.~\cite{bmvc2015-srinivas-data_free_prune} proposed merging similar filters into one. Mariet et al.~\cite{iclr2016-mariet-diversity-networks} proposed merging filters in the FC layers with similar output activations into one. Unfortunately, these coarse-grained pruning approaches tend to have lower compression ratios than fine-grained pruning for the same accuracy.

\textbf{\textit{Previous work directly targets reducing the model size.}} However, as discussed in Sec.~\ref{sec:introduction}, the number of weights alone does not dictate the energy consumption. Hence, the energy consumption of the pruned CNNs in the previous work is not minimized.

To address issues highlighted above, \textbf{\textit{we propose a new fine-grained pruning algorithm that specifically targets energy-efficiency}}. It utilizes the estimated energy provided by the methodology described in Sec.~\ref{sec:energy_estimation} to guide the proposed pruning algorithm to aggressively prune the layers with the highest energy consumption with marginal impact on accuracy. Moreover, the pruning algorithm considers the joint influence of weights on the final output feature maps, thus enabling both a higher compression ratio and a larger energy reduction. The combination of these two approaches results in CNNs that are more energy-efficient and compact than previously proposed approaches.

The proposed energy-efficient pruning algorithm can be combined with other techniques to further reduce the energy consumption, such as bitwidth reduction of weights or feature maps~\cite{eccv2016-rastegari-xnor_net, nips2015-courbariaux-binaryconnect, cvpr2016-wu-quant-cnn}, weight sharing and Huffman coding~\cite{iclr2016-han-deep_comp}, student-teacher learning~\cite{nips2014-ba-teacher-student}, filter decomposition~\cite{iclr2016-kim-comp-cnn-mobile, cvpr2015-foroosh-sparse-cnn} and pruning feature maps~\cite{isca2016-reagen-minerva}.

%% file: 4_energy_aware_pruning.tex
\section{Energy-Aware Pruning}
\label{sec:energy_aware_pruning}

\begin{figure}
\begin{center}
   \includegraphics[width=1.0\linewidth]{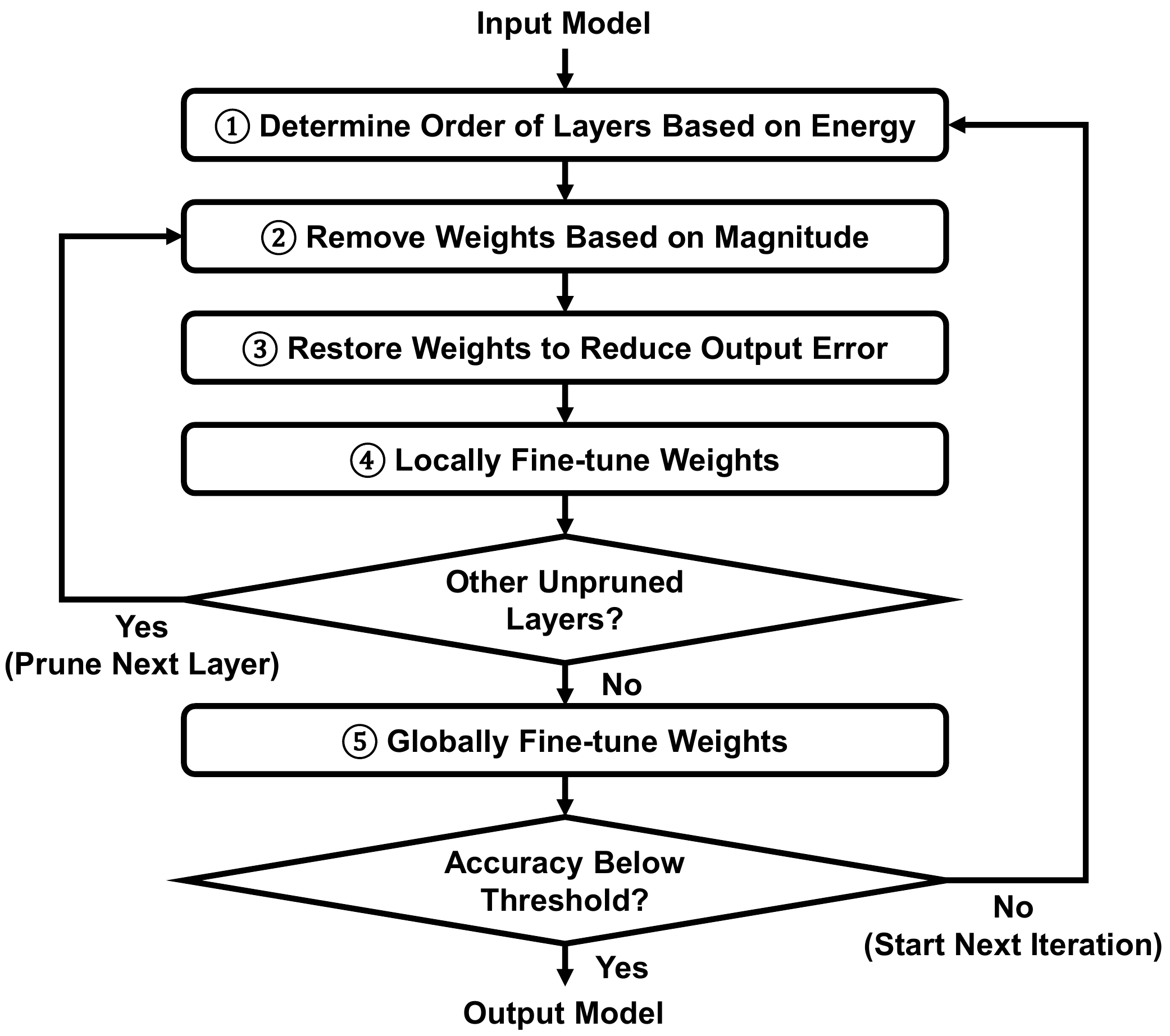}
\end{center}
   \vspace{-5pt}
   \caption{Flow of energy-aware pruning.}
   \vspace{-10pt}
\label{fig:AlgorithmFlow}
\end{figure}

Our goal is to reduce the energy consumption of a given CNN by sparsifying the filters without significant impact on the network accuracy. The key steps in the proposed energy-aware pruning are shown in Fig.~\ref{fig:AlgorithmFlow}, where the input is a CNN model and the output is a sparser CNN model with lower energy consumption.

In \textbf{Step 1}, the pruning order of the layers is determined based on the energy as described in Sec.~\ref{sec:energy_estimation}. Step 2, 3 and 4 removes, restores and locally fine-tunes weights, respectively, for one layer in the network; this inner loop is repeated for each layer in the network. \emph{Pruning} and \emph{restoring} weights involve choosing weights, while \emph{locally fine-tuning} weights involves changing the values of the weights, all while minimizing the output feature map error. In \textbf{Step 2}, a simple magnitude-based pruning method is used to quickly remove the weights above the target compression ratio (\eg, if the target compression ratio is 30\%, 35\% of the weights are removed in this step). The number of extra weights removed is determined empirically. In \textbf{Step 3}, the correlated weights that have the greatest impact on reducing the output error are restored to their original non-zero values to reach the target compression ratio (\eg, restore 5\% of weights). In \textbf{Step 4}, the preserved weights are locally fine-tuned with a closed-form least-square solution to further decrease the output feature map error. Each of these steps are described in detail in Sec.~\ref{sec:step1_order} to Sec.~\ref{sec:step4_finetune_local}.

Once each individual layer has been pruned using Step 2 to 4, \textbf{Step 5} performs global fine-tuning of weights across the entire network using back-propagation as described in Sec.~\ref{sec:step5_finetune_global}. All these steps are iteratively performed until the final network can no longer maintain a given accuracy, \eg, 1\% accuracy loss.

Compared to the previous magnitude-based pruning approaches~\cite{book-hertz-theory-neural-computation, nips2015-han-learn_conn, iclr2016-han-deep_comp, arxiv2016-jin-skinny_net, nips2016-guo-dynamic_surgery}, the main difference of this work is the introduction of Step 1, 3, and 4. Step 1 enables pruning to minimize the energy consumption. Step 3 and 4 increase the compression ratio and reduce the energy consumption.

\subsection{Determine Order of Layers Based on Energy}
\label{sec:step1_order}

As more layers are pruned, it becomes increasingly difficult to remove weights because the accuracy approaches the given accuracy threshold. Accordingly, layers that are pruned early on tend to have higher compression ratios than the layers that follow.  Thus, in order to maximize the overall energy reduction, we prune the layers that consume the most energy first.  Specifically, we use the energy estimation from Sec.~\ref{sec:energy_estimation} and determine the pruning order of layers based on their energy consumption.  As a result, the layers that consume the most energy achieve higher compression ratios and energy reduction. At the beginning of each outer loop iteration in Fig.~\ref{fig:AlgorithmFlow}, the new pruning order is redetermined according to the new energy estimation of each layer.

\subsection{Remove Weights Based on Magnitude}
\label{sec:step2_mag}

For a FC layer, $Y_{i}~\in~\mathbb{R}^{k \times 1}$ is the $i^{th}$ output feature map across $k$ images and is computed from
\begin{equation}
\footnotesize
Y_{i}=X_{i}A_{i}+B_{i}\mathds{1},
\label{eq:FC_Layer}
\end{equation}
where $A_{i}~\in~\mathbb{R}^{m \times 1}$ is the $i^{th}$ filter among all $n$ filters ($A~\in~\mathbb{R}^{m \times n}$) with $m$ weights, and $X_{i}~\in~\mathbb{R}^{k \times m}$ denotes the corresponding $k$ input feature maps, $B_{i}~\in~\mathbb{R}$ is the $i^{th}$ bias, and $\mathds{1}~\in~\mathbb{R}^{k \times 1}$ is a vector where all entries are one. For a CONV layer, we can convert the convolutional operation into a matrix multiplication operation, by converting the input feature maps into a Toeplitz matrix, and compute the output feature maps with a similar equation as Eq.(1).

To sparsify the filters without impacting the accuracy, the simplest method is pruning weights with magnitudes smaller than a threshold, which is referred to as magnitude-based pruning~\cite{book-hertz-theory-neural-computation, nips2015-han-learn_conn, iclr2016-han-deep_comp, arxiv2016-jin-skinny_net, nips2016-guo-dynamic_surgery}. The advantage of this approach is that it is fast, and works well when a few weights are removed, and thus the correlation between weights only has a minor impact on the output. However, as more weights are pruned, this method introduces a large output error as the correlation between weights becomes more critical. For example, if most of the small-magnitude weights are negative, the output error will become large once many of these small negative weights are removed using the magnitude-based pruning. In this case, it would be desirable to remove a large positive weight to compensate for the introduced error instead of removing more smaller negative weights. Thus, we only use magnitude-based pruning for fast initial pruning of each layer. We then introduce additional steps that account for the correlation between weights to reduce the output error due to the magnitude-based pruning.

\subsection{Restore Weights to Reduce Output Error}
\label{sec:step3_restore}


It is the error in the output feature maps, and not the filters, that affects the overall network accuracy.  Therefore, we focus on minimizing the error of the output feature maps instead of that of the filters. To achieve this, we model the problem as the following $\ell0$-minimization problem:
\begin{equation}
\footnotesize
\begin{aligned}
\tilde{A}_{i}=\argmin_{\hat{A}_{i}} \left \| \hat{Y}_{i}-X_{i}\hat{A}_{i} \right \|_{p}^p,
\\ \text{subject to} \left \| \hat{A} \right \|_{0} \leqslant {q},\quad i = 1,...,n,
\end{aligned}
\label{eq:L0_Minimization}
\end{equation}
where $\hat{Y}_{i}$ denotes $Y_{i}-B_{i}\mathds{1}$, $\left \| \cdot \right \|_{p}$ is the $p$-norm, and $q$ is the number of non-zero weights we want to retain in all filters. $p$ can be set to 1 or 2, and we use 1. Unfortunately, solving this $\ell0$-minimization problem is NP-hard. Therefore, a greedy algorithm is proposed to approximate it. 

The algorithm starts from pruned filters $\breve{A}~\in~\mathbb{R}^{m \times n}$, obtained from the magnitude-based pruning in Step 2. These filters are pruned at a higher compression ratio than the target compression ratio. Each filter $A_{i}$ has the corresponding support $S_{i}$, where $S_{i}$ is a set of the indices of non-zero weights in the filter. It then iteratively restores weights until the number of non-zero weights is equal to $q$, which reflects the target compression ratio.

The residual of each filter, which indicates the current output feature map difference we need to minimize, is initialized as $\hat{Y}_{i}-X_{i}\breve{A}_{i}$. In each iteration, out of the weights not in the support of a given filter $S_{i}$, we select the weight that reduces the $\ell1$-norm of the corresponding residual the most, and add it to the support $S_{i}$. The residual then is updated by taking this new weight into account.


We restore weights from the filter with the largest residual in each iteration. This prevents the algorithm from restoring weights in filters with small residuals, which will likely have less effect on the overall output feature map error. This could occur if the weights were selected based solely on the largest $\ell$1-norm improvement for any filter.

To speed up this restoration process, we restore multiple weights within a given filter in each iteration.  The $g$ weights with the top-$g$ maximum $\ell1$-norm improvement are chosen. As a result, we reduce the frequency of computing residual improvement for each weight, which takes a significant amount of time. We adopt $g$ equal to 2 in our experiments, but a higher $g$ can be used. 

\subsection{Locally Fine-tune Weights}
\label{sec:step4_finetune_local}

The previous two steps select a subset of weights to preserve, but do not change the values of the weights. In this step, we perform the least-square optimization on each filter to change the values of their weights to further reduce the output error and restore the network accuracy: 
\begin{equation}
\footnotesize
\bar{A}_{i,S_{i}}=\argmin_{\hat{A}_{i,S_{i}}} \left \| \hat{Y}_{i}-X_{i,S_{i}}\hat{A}_{i,S_{i}} \right \|_{2}^2,\quad \bar{A}_{i,S_{i}^C}=0,
\label{eq:LeastSquareProblem}
\end{equation}
where the subscript $S_{i}$ means choosing the non-pruned weights from the $i^{th}$ filter and the corresponding columns from $X_{i}$. The least-square problem has a closed-form solution, which can be efficiently solved.

\subsection{Globally Fine-tune Weights}
\label{sec:step5_finetune_global}
After all the layers are pruned, we fine-tune the whole network using back-propagation with the pruned weights fixed at zero. This step can be used to globally fine-tune the weights to achieve a higher accuracy. Fine-tuning the whole network is time-consuming and requires careful tuning of several hyper-parameters. In addition, back-propagation can only restore the accuracy within certain accuracy loss. However, since we first locally fine-tune weights, part of the accuracy has already been restored, which enables more weights to be pruned under a given accuracy loss tolerance. As a result, we increase the compression ratio in each iteration, reducing the total number of globally fine-tuning iterations and the corresponding time.


%% file: 5_experiment_results.tex
\section{Experiment Results}
\label{sec:experiment_results}


\subsection{Pruning Method Evaluation}

We evaluate our energy-aware pruning on AlexNet~\cite{nips2012-krizhevsky-alexnet}, GoogLeNet v1~\cite{cvpr2015-szegedy-googlenet} and SqueezeNet v1~\cite{arxiv2016-iandola-squeezenet} and compare it with the state-of-the-art magnitude-based pruning method with the publicly available models~\cite{iclr2016-han-deep_comp}.\footnote{The proposed energy-aware pruning can be easily combined with other techniques in \cite{iclr2016-han-deep_comp}, such as weight sharing and Huffman coding.} The accuracy and the energy consumption are measured on the ImageNet ILSVRC 2014 dataset~\cite{ijcv2015-russakovsky-ilsvrc}. Since the energy-aware pruning method relies on the output feature maps, we use the training images for both pruning and fine-tuning. All accuracy numbers are measured on the validation images. To estimate the energy consumption with the proposed methodology in Sec.~\ref{sec:energy_estimation}, we assume all values are represented with 16-bit precision, except where otherwise specified, to fairly compare the energy consumption of networks. The hardware parameters used are similar to ~\cite{isscc2016-chen-eyeriss_chip}.

\begin{table*}[t]
\small
\centering
\begin{threeparttable}[b]
\caption{Performance metrics of various dense and pruned models.}
\label{tab:NetworkPruningSummary}
\begin{tabular}{rl|c|rr|rr|rr}
\multicolumn{2}{c|}{\textbf{Model}}                                  			& \textbf{\begin{tabular}[c]{@{}c@{}} Top-5\\Accuracy\end{tabular}} & \multicolumn{2}{c|}{\textbf{\begin{tabular}[c]{@{}c@{}}\# of Non-zero\\ Weights ($\times10^6$)\end{tabular}}} 	 	& \multicolumn{2}{c|}{\textbf{\begin{tabular}[c]{@{}c@{}}\# of Non-skipped\\ MACs ($\times10^8$)\tnote{1}\end{tabular}}} 	&  \multicolumn{2}{c}{\textbf{\begin{tabular}[c]{@{}c@{}}Normalized \\ Energy ($\times10^9$)\tnote{1,2}\end{tabular}}} 	\\ \hline
AlexNet 			& (Original)               											& 80.43\%                                                           													& 60.95 	&  (100\%)                         																																							& 3.71 	& (100\%)            																																					& 3.97 	& (100\%)                      																																										\\ 
AlexNet 			& (\cite{iclr2016-han-deep_comp})              & 80.37\%                                                           													& 6.79 		& (11\%)                         																																								& 1.79 	& (48\%)             																																					& 1.85 	& (47\%)                    																																											\\ 
AlexNet 			& (Energy-Aware Pruning)             					& 79.56\%                                                          	 												& 5.73 		& (9\%)                         																																									& 0.56 	& (15\%)            																																					& 1.06 	& (27\%)                    																																											\\ \hline
GoogLeNet 		& (Original)         													& 88.26\%                                                           													& 6.99 		& (100\%)                          																																							& 7.41 	& (100\%)               																																				& 7.63 	& (100\%)                     		 																																								\\ 
GoogLeNet 		& (Energy-Aware Pruning)        						& 87.28\%                                                           													& 2.37 		& (34\%)                        																																								& 2.16 	& (29\%)             																																					& 4.76 	& (62\%)                   																																											\\ \hline
SqueezeNet 	& (Original)       													& 80.61\%                                                           													& 1.24 		& (100\%)                          																																							& 4.51 	& (100\%)               																																				& 5.28 	& (100\%)                      																																										\\ 
SqueezeNet 	& (\cite{iclr2016-han-deep_comp})       		& 81.47\%                                                           													& 0.42 		& (33\%)                        																																								& 3.30 	& (73\%)             																																					& 4.61 	& (87\%)                    																																											\\ 
SqueezeNet 	& (Energy-Aware Pruning)      						& 80.47\%                                                           													& 0.35 		& (28\%)                       																																								& 1.93 	& (43\%)             																																					& 3.99 	& (76\%)                    																																											\\ 
\end{tabular}
\begin{tablenotes}
    \item[1] Per image.
    \item[2] The unit of energy is normalized in terms of the energy for a MAC operation (\ie, $10^2$ = energy of 100 MACs). 
\end{tablenotes}
\end{threeparttable}
\end{table*}

\begin{table}[t]
\small
\centering
\begin{threeparttable}[b]
\caption{Compression ratio\tnote{1} of each layer in AlexNet.}
\label{tab:PruneAlexNet}
\begin{tabular}{c|c|c|c|c|c}
\textbf{}                                                         & \textbf{\cite{iclr2016-han-deep_comp}}   & \multicolumn{4}{c}{\textbf{This Work}} 
\\ \hline 
\textbf{\begin{tabular}[c]{@{}c@{}}\# of \\ Classes\end{tabular}} & \textbf{1000} & \textbf{1000} & \textbf{100} & \textbf{\begin{tabular}[c]{@{}c@{}}10\\ (Random)\end{tabular}} & \textbf{\begin{tabular}[c]{@{}c@{}}10\\ (Dog)\end{tabular}} \\ \hline
CONV1                                                             & 16\%          & 83\%          & 86\%         & 89\%                                                           & 89\%                                                        \\ 
CONV2                                                             & 62\%          & 92\%          & 97\%         & 97\%                                                           & 96\%                                                        \\ 
CONV3                                                             & 65\%          & 91\%          & 97\%         & 98\%                                                           & 97\%                                                        \\ 
CONV4                                                             & 63\%          & 81\%          & 88\%         & 97\%                                                           & 95\%                                                        \\ 
CONV5                                                             & 63\%          & 74\%          & 79\%         & 98\%                                                           & 98\%                                                        \\ 
FC1                                                               & 91\%          & 92\%          & 93\%         & $\sim$100\%                                                    & $\sim$100\%                                                 \\ 
FC2                                                               & 91\%          & 91\%          & 94\%         & $\sim$100\%                                                    & $\sim$100\%                                                 \\ 
FC3                                                               & 74\%          & 78\%          & 78\%         & $\sim$100\%                                                    & $\sim$100\%                                                 \\ 
\end{tabular}
\begin{tablenotes}
    \item[1] The number of removed weights divided by the number of total weights. The higher, the better.
\end{tablenotes}
\vspace{-11pt}
\end{threeparttable}
\end{table}

Table~\ref{tab:NetworkPruningSummary} summarizes the results.\footnote{We use the models provided by MatConvNet~\cite{mm2015-vedaldi-matconvnet} or converted from Caffe~\cite{arxiv2014-jia-caffe} or Torch~\cite{nips2011-collobert-torch7}, so the accuracies may be slightly different from that reported by other works.} The batch size is 44 for AlexNet and 48 for other two networks. All the energy-aware pruned networks have less than 1\% accuracy loss with respect to the other corresponding networks. For AlexNet and SqueezeNet, our method achieves better results in all metrics (\ie, number of weights, number of MACs, and energy consumption) than the magnitude-based pruning~\cite{iclr2016-han-deep_comp}. For example, the number of MACs is reduced by another 3.2$\times$ and the estimated energy is reduced by another 1.7$\times$ with a 15\% smaller model size on AlexNet. Table~\ref{tab:PruneAlexNet} shows a comparison of the energy-aware pruning and the magnitude-based pruning across each layer; our method gives a higher compression ratio for all layers, especially for CONV1 to CONV3, which consume most of the energy.

Our approach is also effective on compact models. For example, on GoogLeNet, the achieved reduction factor is 2.9$\times$ for the model size, 3.4$\times$ for the number of MACs and 1.6$\times$ for the estimated energy consumption.

\subsection{Energy Consumption Analysis}
\label{subsec:EnergyConsumptionAnalysis}

\begin{figure*}[t]
\vspace{-5pt}
\begin{center}
   \includegraphics[width=0.72\linewidth]{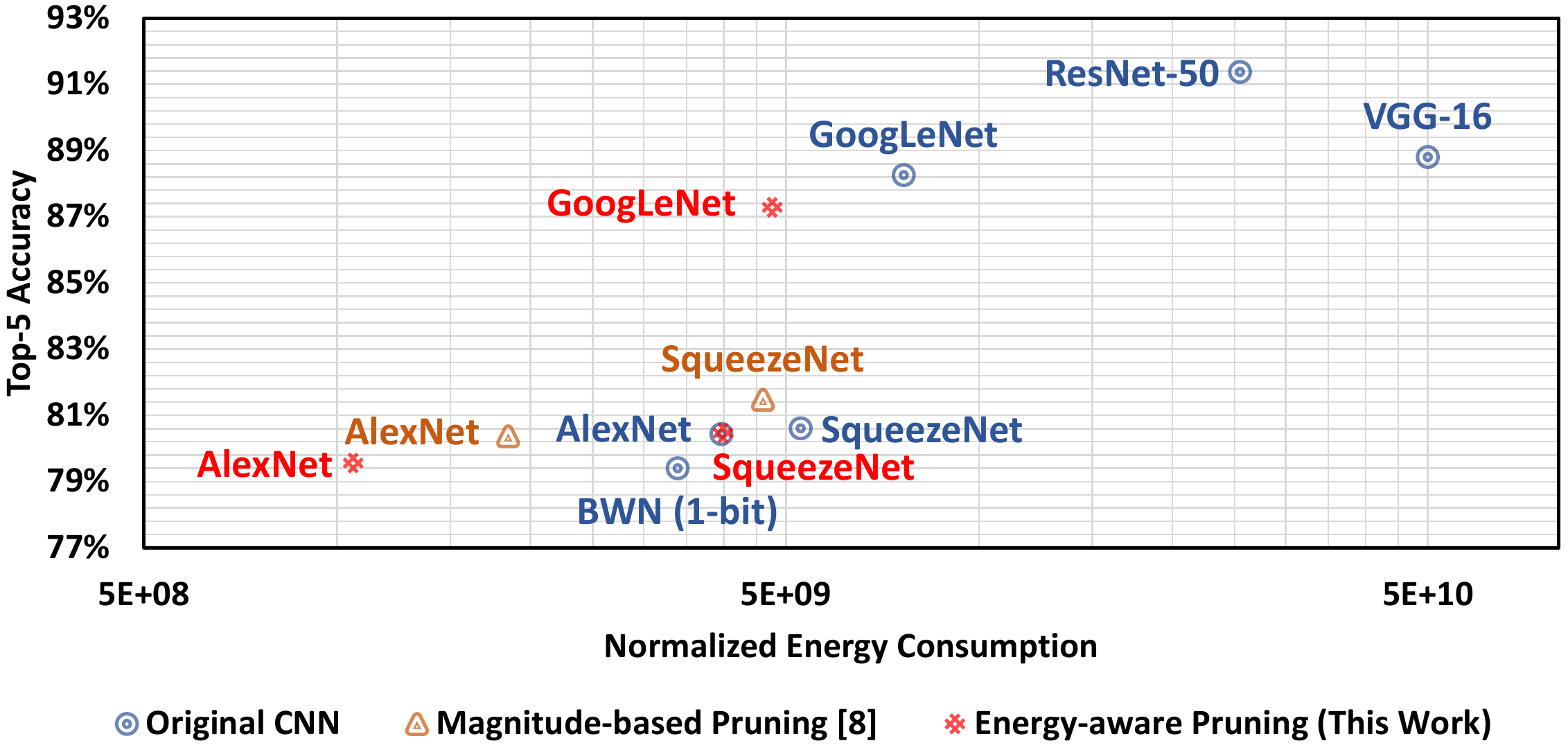}
\end{center}
   \vspace{-10pt}
   \caption{Accuracy versus energy trade-off of popular CNN models. Models pruned with the energy-aware pruning provide a better accuracy versus energy trade-off (steeper slope).}
   \vspace{-12pt}
\label{fig:CompreModels}
\end{figure*}

\begin{figure}[t]
\vspace{-11pt}
\begin{center}
   \includegraphics[width=0.69\linewidth]{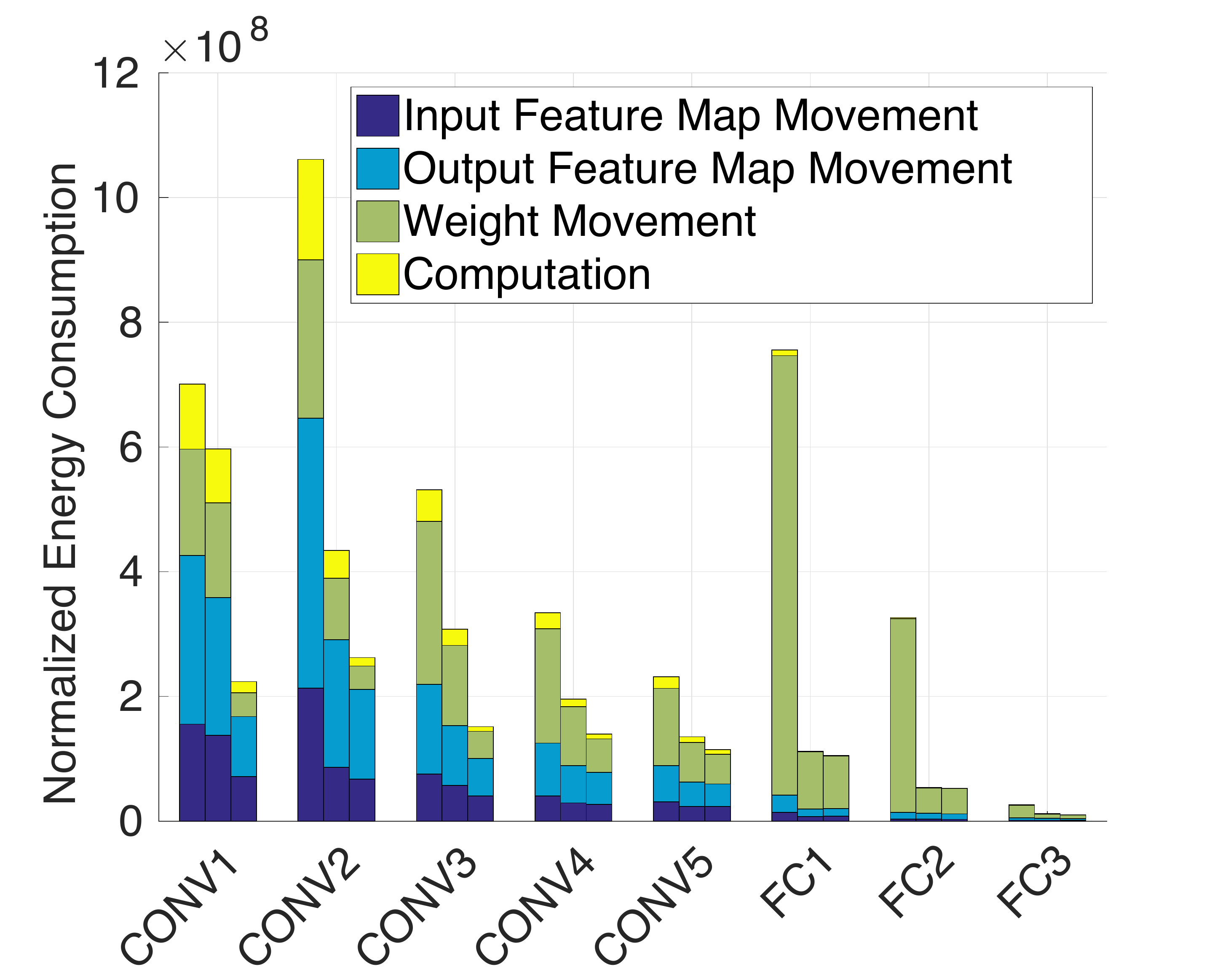}
\end{center}
\vspace{-11pt}
   \caption{Energy consumption breakdown of different AlexNets in terms of the computation and the data movement of input feature maps, output feature maps and filter weights. From left to right: original AlexNet, AlexNet pruned by \cite{iclr2016-han-deep_comp}, AlexNet pruned by the proposed energy-aware pruning.}
   \vspace{-10pt}
\label{fig:EnergyBreakdown_Prune}
\end{figure}


We also evaluate the energy consumption of popular CNNs. In Fig.~\ref{fig:CompreModels}, we summarize the estimated energy consumption of CNNs relative to their top-5 accuracy. The results reveal the following key observations:

\begin{list}{\labelitemi}{\leftmargin=1em}
  \setlength{\topmargin}{0pt}
  \setlength{\itemsep}{0em}
  \setlength{\parskip}{0pt}
  \setlength{\parsep}{0pt}

\item \textbf{Convolutional layers consume more energy than fully-connected layers.} Fig.~\ref{fig:EnergyBreakdown_Prune} shows the energy breakdown of the original AlexNet and two pruned AlexNet models. Although most of the weights are in the FC layers, CONV layers account for most of the energy consumption. For example, in the original AlexNet, the CONV layers contain 3.8\% of the total weights, but consume 72.6\% of the total energy. There are two reasons for this: (1) In CONV layers, the energy consumption of the input and output feature maps is much higher than that of FC layers. Compared to FC layers, CONV layers require a larger number of MACs, which involves loading inputs from memory and writing the outputs to memory. Accordingly, a large number of MACs leads to a large amount of weight and feature map movement and hence high energy consumption; (2) The energy consumption of weights for all CONV layers is similar to that of all FC layers. While CONV layers have fewer weights than FC layers, each weight in CONV layers is used more frequently than that in FC layers; this is the reason why the number of weights is not a good metric for energy consumption -- different weights consume different amounts of energy.  Accordingly, pruning a weight from CONV layers contributes more to energy reduction than pruning a weight from FC layers. In addition, as a network goes deeper, \eg, ResNet~\cite{cvpr2016-he-resnet}, CONV layers dominate both the energy consumption and the model size.  The energy-aware pruning prunes CONV layers effectively, which significantly reduces energy consumption.

\item \textbf{Deeper CNNs with fewer weights do not necessarily consume less energy than shallower CNNs with more weights.}
One network design strategy for reducing the size of a network without sacrificing the accuracy is to make a network thinner but deeper. However, does this mean the energy consumption is also reduced? Table~\ref{tab:NetworkPruningSummary} shows that a network architecture having a smaller model size does not necessarily have lower energy consumption. For instance, SqueezeNet is a compact model and a good fit for memory-limited applications; it is thinner and deeper than AlexNet and achieves a similar accuracy with 50$\times$ size reduction, but consumes 33\% more energy. The increase in energy is due to the fact that SqueezeNet uses more CONV layers and the size of the feature maps can only be greatly reduced in the final few layers to preserve the accuracy. Hence, the newly added CONV layers involve a large amount of computation and data movement, resulting in higher energy consumption.

\item \textbf{Reducing the number of weights can provide lower energy consumption than reducing the bitwidth of weights.} From Fig.~\ref{fig:CompreModels}, the AlexNet pruned by the proposed method consumes less energy than BWN~\cite{eccv2016-rastegari-xnor_net}. BWN uses an AlexNet-like architecture with binarized weights, which \emph{only} reduces the weight-related and computation-related energy consumption. However, pruning reduces the energy of \emph{both} weight and feature map movement, as well as computation. In addition, the weights in CONV1 and FC3 of BWN are not binarized to preserve the accuracy; thus BWN does not reduce the energy consumption of CONV1 and FC3. Moreover, to compensate for the accuracy loss of binarizing the weights, CONV2, CONV4 and CONV5 layers in BWN use 2$\times$ the number of weights in the corresponding layers of the original AlexNet, which increases the energy consumption.

\item \textbf{A lower number of MACs does not necessarily lead to lower energy consumption.} For example, the pruned GoogleNet has a fewer MACs but consumes more energy than the SqueezeNet pruned by \cite{iclr2016-han-deep_comp}. That is because they have different data reuse, which is determined by the shape configurations, as discussed in Sec.~\ref{subsec:energy_estimation_background}.


\end{list}

From Fig.~\ref{fig:CompreModels}, we also observe that the energy consumption scales exponentially with linear increase in accuracy. For instance, GoogLeNet consumes 2$\times$ energy of AlexNet for 8\% accuracy improvement, and ResNet-50 consumes 3.3$\times$ energy of GoogLeNet for 3\% accuracy improvement. 

In summary, the model size (\ie, the number of weights $\times$ the bitwidth) and the number of MACs do not directly reflect the energy consumption of a layer or a network. There are other factors like the data movement of the feature maps, which are often overlooked. Therefore, with the proposed energy estimation methodology, researchers can have a clearer view of CNNs and more effectively design low-energy-consumption networks.

\subsection{Number of Target Class Reduction}

\begin{figure}[t]
    \centering
    \subfloat[\# of weights]
    {
        \includegraphics[width=0.14\textwidth]{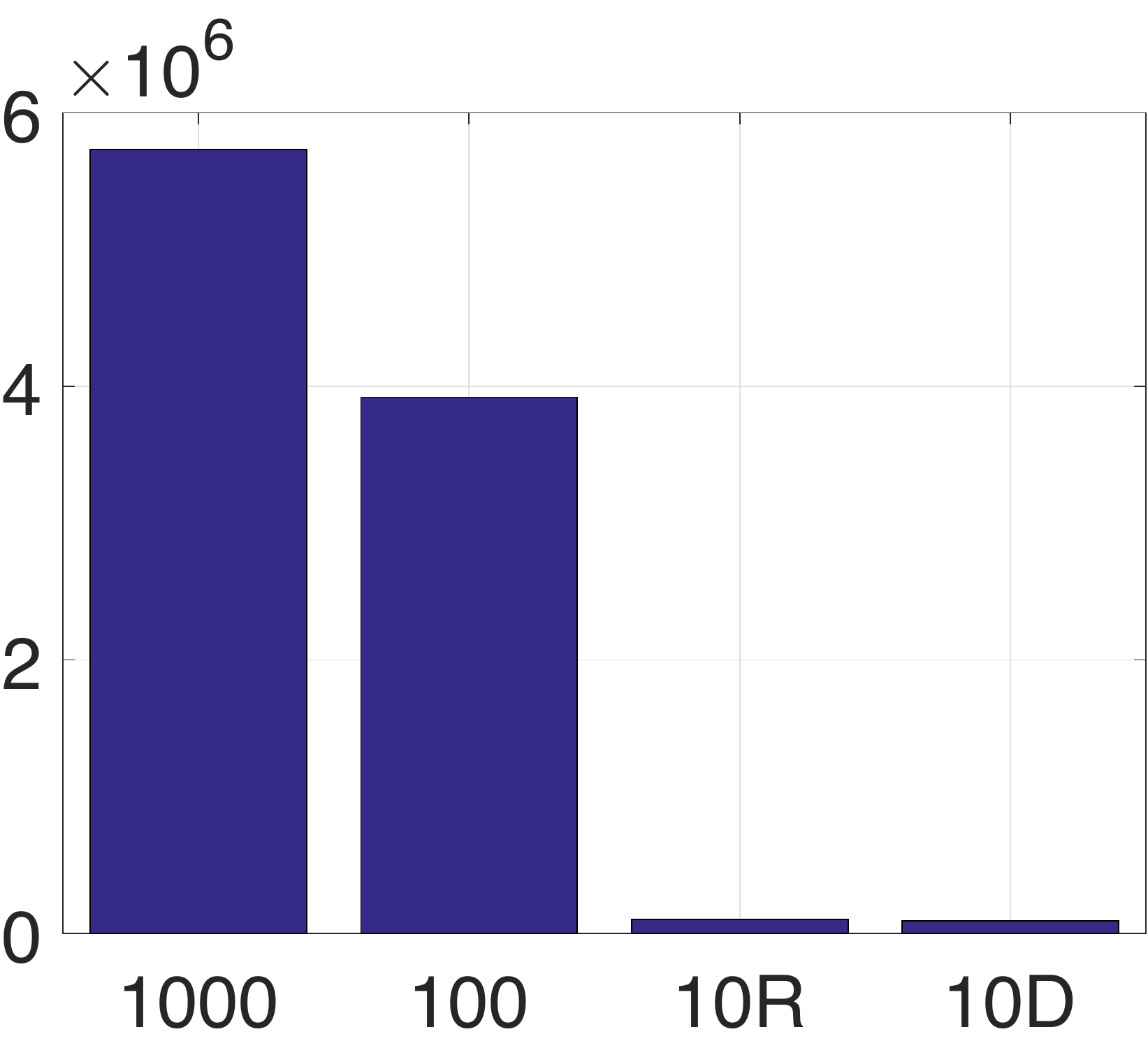}
    }
    \subfloat[\# of MACs]
    {
        \includegraphics[width=0.14\textwidth]{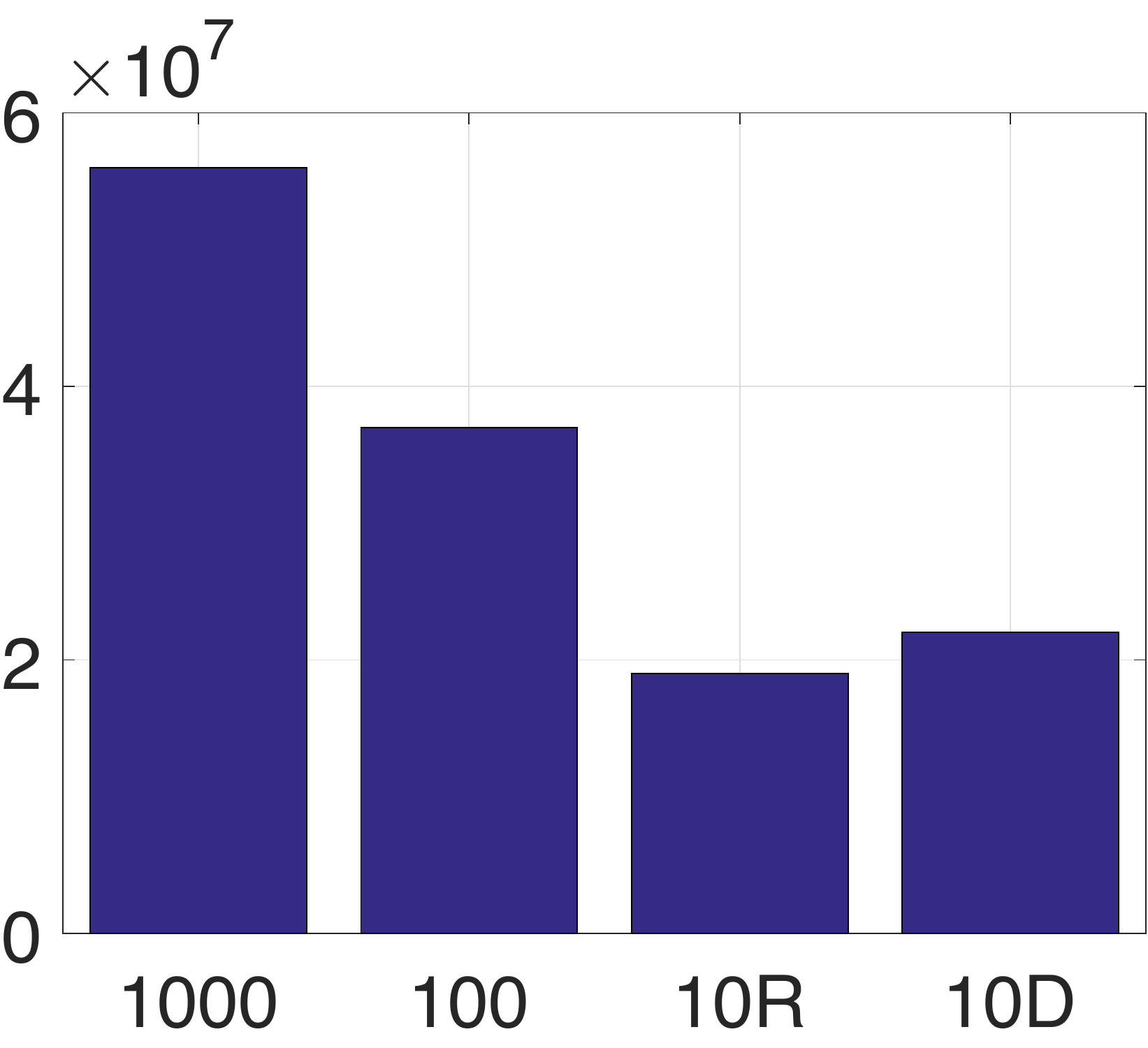}
    }
    \subfloat[Estimated energy]
    {
        \includegraphics[width=0.14\textwidth]{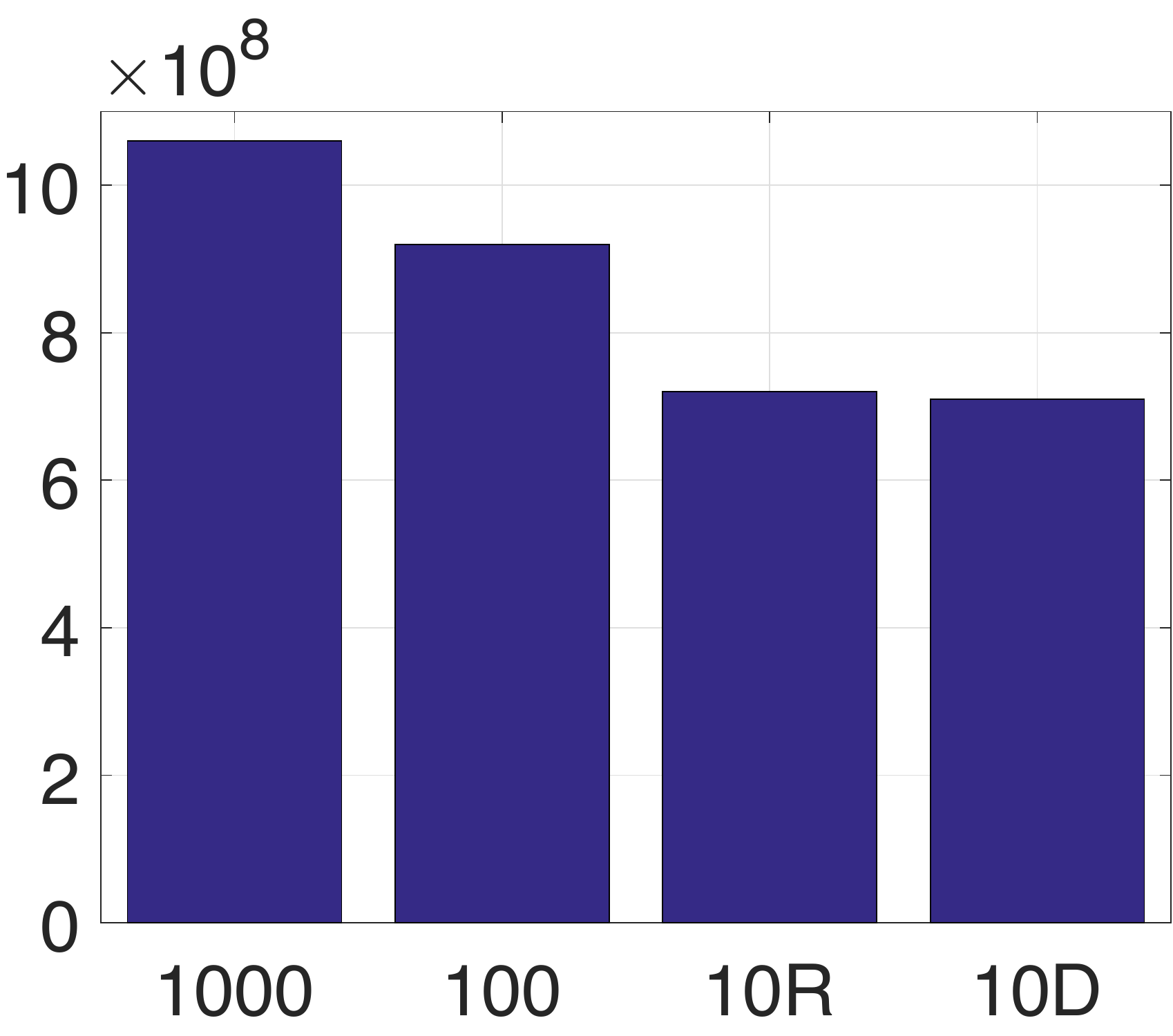}
    }
    \caption{The impact of reducing the number of target classes on the three metrics. The x-axis is the number of target classes. 10R and 10D denote the 10-random-class model and the 10-dog-class model, respectively.}
        \vspace{-15pt}
    \label{fig:ReduceNumberOfClassSummary}
\end{figure}

\begin{figure}[t]
    \centering
    \subfloat[Input feature map]
    {
        \includegraphics[width=0.14\textwidth]{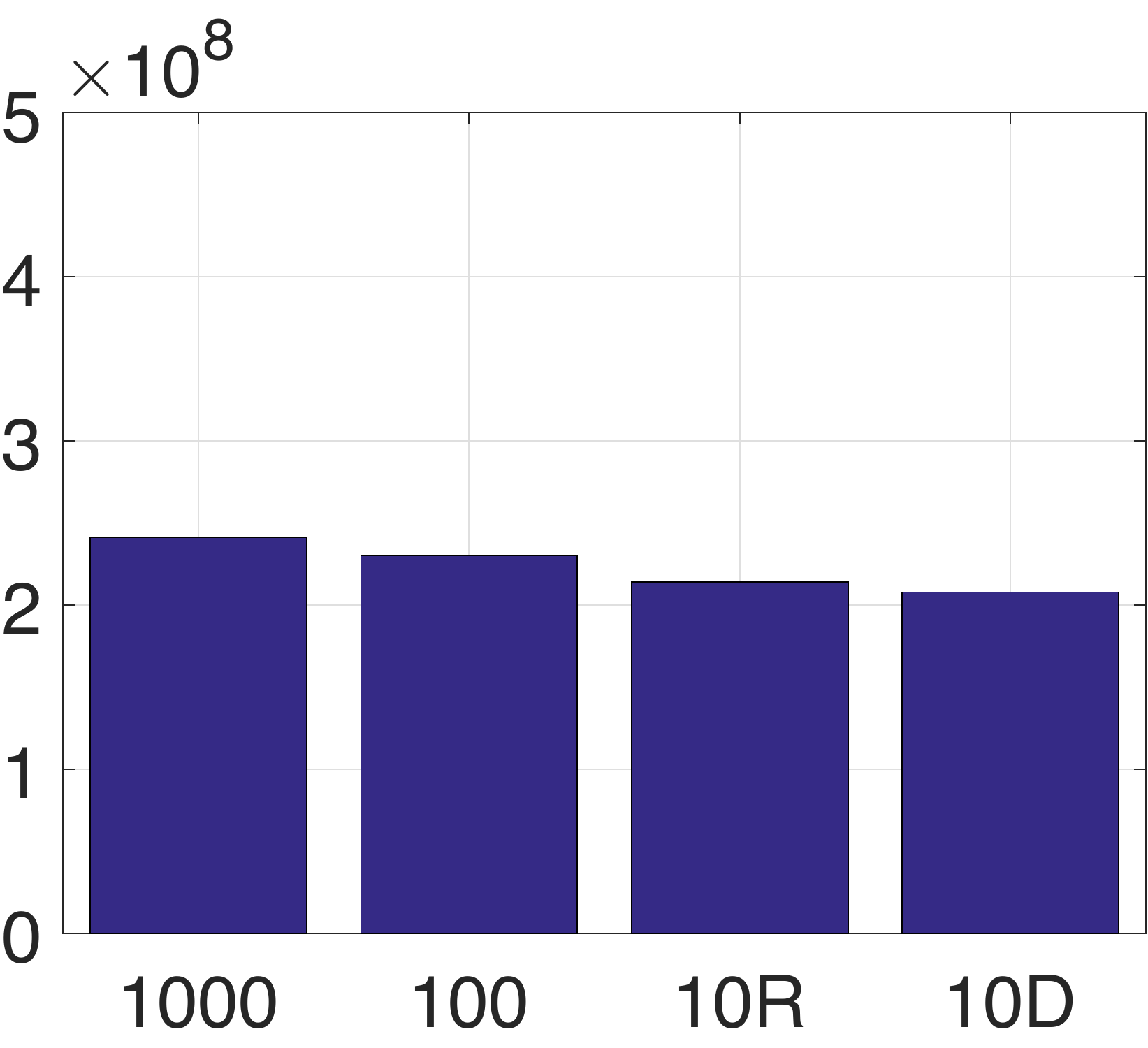}
    }
    \subfloat[Output feature map]
    {
        \includegraphics[width=0.14\textwidth]{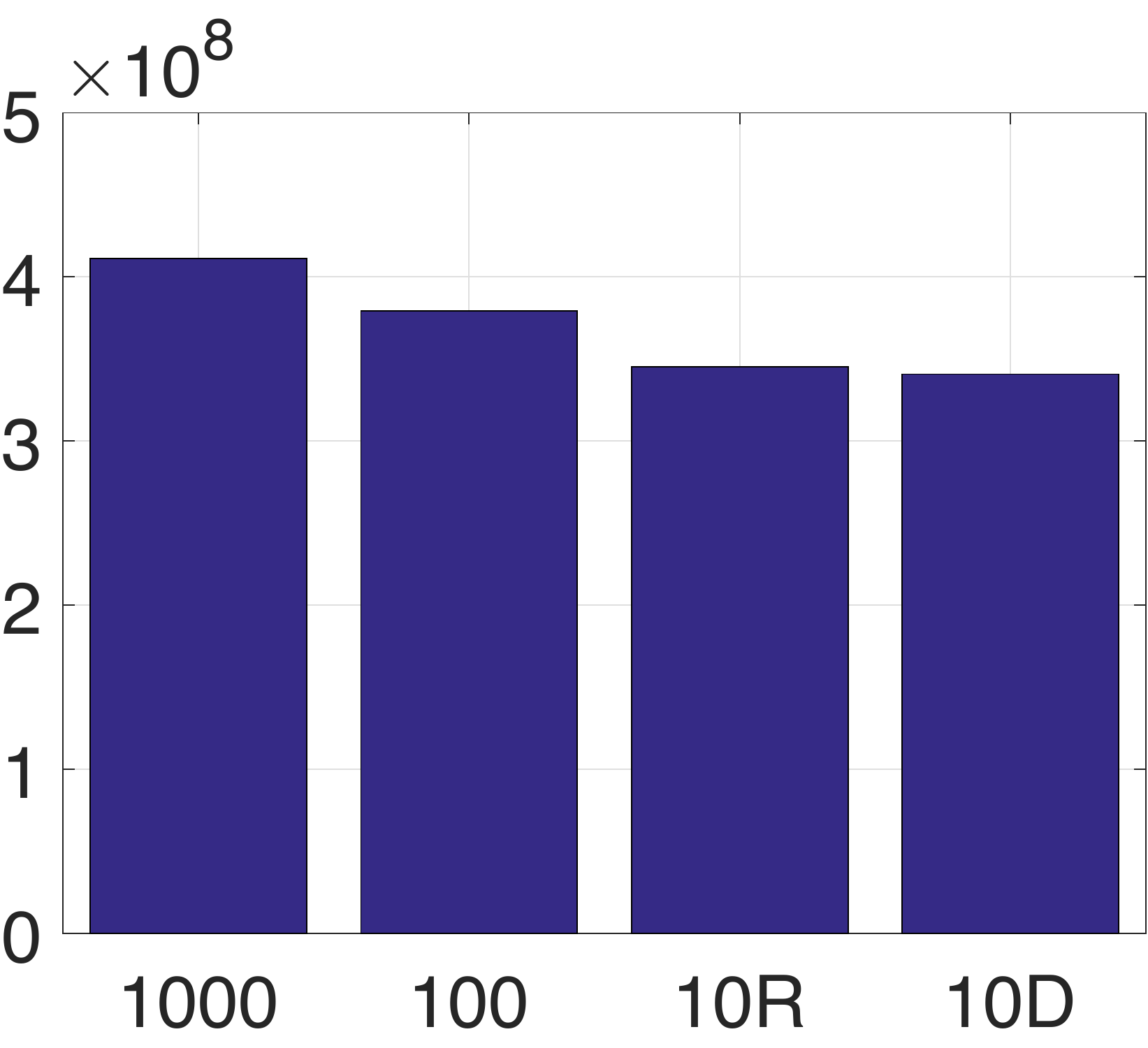}
    }
    \subfloat[Weight]
    {
        \includegraphics[width=0.14\textwidth]{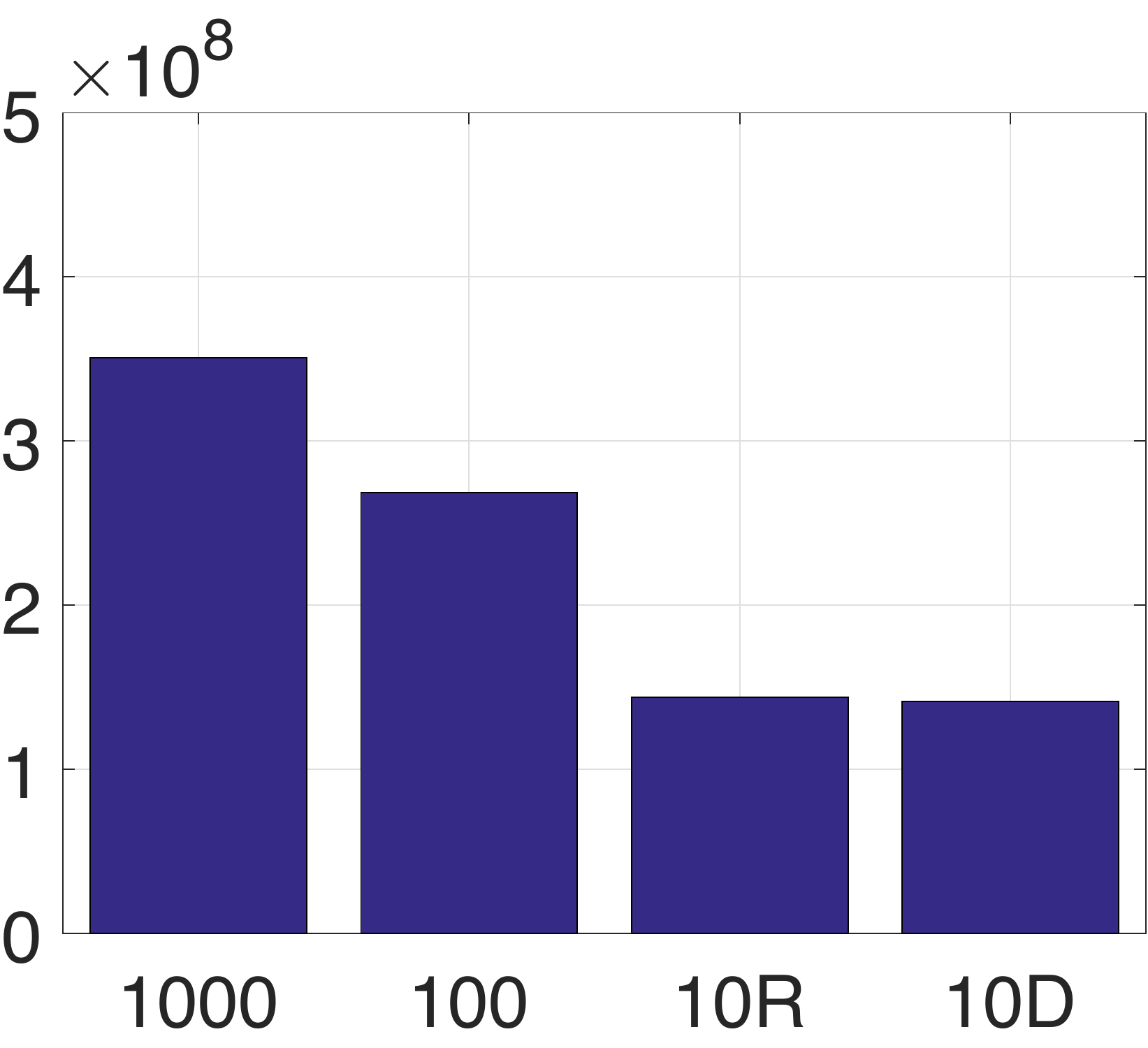}
    }
    \caption{The energy breakdown of models with different numbers of target classes.}
    \vspace{-10pt}
    \label{fig:EnergyBreakdown_ReduceClass}
\end{figure}

In many applications, the number of classes can be significantly fewer than 1000. We study the influence of reducing the number of target classes by pruning weights on the three metrics. AlexNet is used as the starting point. The number of target classes is reduced from 1000 to 100 to 10. The target classes of the 100-class model and one of the 10-class models are randomly picked, and that of another 10-class model are different dog breeds. These models are pruned with less than $1\%$ top-5 accuracy loss for the 100-class model and less than $1\%$ top-1 accuracy loss for the two 10-class models.

Fig.~\ref{fig:ReduceNumberOfClassSummary} shows that as the number of target classes reduces, the number of weights and MACs and the estimated energy consumption decrease. However, they reduce at different rates with the model size dropping the fastest, followed by the number of MACs the second, and the estimated energy reduces the slowest.

According to Table~\ref{tab:PruneAlexNet}, for the 10-class models, almost all the weights in the FC layers are pruned, which leads to a very small model size. Because the FC layers work as classifiers, most of the weights that are responsible for classifying the removed classes are pruned. The higher-level CONV layers, such as CONV4 and CONV5, which contain filters for extracting more specialized features of objects, are also significantly pruned. CONV1 is pruned less since it extracts basic features that are shared among all classes. As a result, the number of MACs and the energy consumption do not reduce as rapidly as the number of weights. Thus, we hypothesize that the layers closer to the output of a network shrink more rapidly with the number of classes.

As the number of classes reduces, the energy consumption becomes less sensitive to the filter sparsity. From the energy breakdown (Fig.~\ref{fig:EnergyBreakdown_ReduceClass}), the energy consumption of feature maps gradually saturates due to data reuse and the memory hierarchy. For example, each time one input activation is loaded from the DRAM onto the chip, it is used multiple times by several weights. If any one of these weights is not pruned, the activation still needs to be fetched from the DRAM. Moreover, we observe that sometimes the sparsity of feature maps decreases after we reduce the number of target classes, which causes higher energy consumption for moving the feature maps.


Table~\ref{tab:PruneAlexNet} and Fig.~\ref{fig:ReduceNumberOfClassSummary} and \ref{fig:EnergyBreakdown_ReduceClass} show that the compression ratios and the performance of the two 10-class models are similar. Hence, we hypothesize that the pruning performance mainly depends on the number of target classes, and the type of the preserved classes is less influential.

%% file: 6_conclusion.tex
\section{Conclusion}
\label{sec:conclusion}

This work presents an energy-aware pruning algorithm that directly uses the energy consumption of a CNN to guide the pruning process in order to optimize for the best energy-efficiency. The energy of a CNN is estimated by a methodology that models the computation and memory accesses of a CNN and uses energy numbers extrapolated from actual hardware measurements. It enables more accurate energy consumption estimation compared to just using the model size or the number of MACs. With the estimated energy for each layer in a CNN model, the algorithm performs layer-by-layer pruning, starting from the layers with the highest energy consumption to the layers with the lowest energy consumption. For pruning each layer, it removes the weights that have the smallest joint impact on the output feature maps. The experiments show that the proposed pruning method reduces the energy consumption of AlexNet and GoogLeNet, by 3.7$\times$ and 1.6$\times$, respectively, compared to their original dense models. The influence of pruning the AlexNet with the number of target classes reduced is explored and discussed. The results show that by reducing the number of target classes, the model size can be greatly reduced but the energy reduction is limited.